\documentclass[a4paper,fleqn]{cas-sc}

\usepackage[numbers]{natbib}
\usepackage{subfig}
\usepackage{balance}
\usepackage{algorithmic}
\usepackage{algorithm}

\def\tsc#1{\csdef{#1}{\textsc{\lowercase{#1}}\xspace}}
\tsc{WGM}
\tsc{QE}

\begin{document}
\let\WriteBookmarks\relax
\def\floatpagepagefraction{1}
\def\textpagefraction{.001}

\shorttitle{}    

\shortauthors{Z. Yin et~al.}  

\title [mode = title]{Dynamic Programming-Based Redundancy Resolution for Path Planning of Redundant Manipulators Considering Breakpoints}

%

\author[1]{Zhihang Yin}[style=chinese]
\fnmark[1]

\affiliation[1]{organization={School of Mathematical Sciences, Zhejiang University},
	addressline={Xihu}, 
	city={Hangzhou},
	postcode={310058}, 
	state={Zhejiang},
	country={China}}

\author[2]{Fa Wu}[style=chinese]
\fnmark[1]

\affiliation[2]{organization={Zhejiang Demetics Medical Technology Co., Ltd.},
	addressline={Xihu}, 
	city={Hangzhou},
	postcode={310012}, 
	state={Zhejiang},
	country={China}}
	
	\author[1]{Ruofan Bian}[style=chinese]
	\fnmark[1]

	
\author[3]{Ziqian Wang}[style=chinese]

\affiliation[3]{organization={Department of Human Development, Teachers College, Columbia University}, 
	city={New York},
	postcode={10027}, 
	state={NY},
	country={United States}}

\author[4]{Jianmin Yang}[style=chinese]

\affiliation[4]{organization={Zhejiang College of Sports},
	city={Hangzhou},
	postcode={311231}, 
	state={Zhejiang},
	country={China}}

\author[5]{Jiyong Tan}[style=chinese, orcid=0000-0001-6356-1743]
\cormark[1]
\ead{scutjy2015@163.com}

\affiliation[5]{organization={Shenzhen Institute for Advanced Study, University of Electronic Science and Technology of China}, 
	city={Shenzhen},
	postcode={610056}, 
	state={Guangdong},
	country={China}}

\author[1]{Dexing Kong}[style=chinese, orcid=0000-0001-9339-8086]
\cormark[1]
\ead{dxkong@zju.edu.cn}

\cortext[cor1]{Corresponding author}
\fntext[fn1]{The two authors contribute equally to this work.}


\begin{abstract}
This paper proposes a redundancy resolution algorithm for a redundant manipulator based on dynamic programming. This algorithm can compute the desired joint angles at each point on a pre-planned discrete path in Cartesian space, while ensuring that the angles, velocities, and accelerations of each joint do not exceed the manipulator's constraints. We obtain the analytical solution to the inverse kinematics problem of the manipulator using a parameterization method, transforming the redundancy resolution problem into an optimization problem of determining the parameters at each path point. The constraints on joint velocity and acceleration serve as constraints for the optimization problem. Then all feasible inverse kinematic solutions for each pose under the joint angle constraints of the manipulator are obtained through parameterization methods, and the globally optimal solution to this problem is obtained through the dynamic programming algorithm. On the other hand, if a feasible joint-space path satisfying the constraints does not exist, the proposed algorithm can compute the minimum number of breakpoints required for the path and partition the path with as few breakpoints as possible to facilitate the manipulator's operation along the path. The algorithm can also determine the optimal selection of breakpoints to minimize the global cost function, rather than simply interrupting when the manipulator is unable to continue operating. The proposed algorithm is tested using a manipulator produced by a certain manufacturer, demonstrating the effectiveness of the algorithm.
\end{abstract}




\maketitle

	\section{Introduction}

In many applications of robotics, such as ultrasound scanning robots\cite{ref1}, massage robots\cite{ref0}, polishing robots\cite{ref01}, and others, it is necessary for the robot to move precisely along a pre-planned path in a three-dimensional workspace. When performing these tasks, traditional 6-DOF manipulators are easily limited by singular points and joint ranges, making it difficult for them to reach certain specific poses along the given path \cite{ref02}. 

On the other hand, redundant manipulators with more independently driven joints than the required poses of the end effector have higher flexibility compared to traditional manipulators. Not only can they effectively accomplish tasks, but they can also optimize the end effector's trajectory according to varying task requirements.

This paper considers the application scenario of automatic ultrasound scanning using a 7-degree-of-freedom redundant manipulator. Equipped with an ultrasound probe at its end effector, the manipulator acquires ultrasound images near the patient's skin. Consequently, the manipulator needs to move along a Cartesian space path specified according to the patient's body surface, aiming to complete the scan as seamlessly as possible to enhance the quality and efficiency of the examination. 

The control of the manipulator is achieved by sending control commands to the control computer at a fixed frequency. Consequently, we discretize the path in Cartesian space. Each sampling point corresponds to an end-effector pose of the manipulator. While many algorithms, such as \cite{ref020,ref0201}, prioritize time minimization as the optimization objective, patient safety and comfort, along with enabling real-time analysis and diagnosis of the ultrasound images by the sonographer, are crucial considerations. Therefore, while planning the manipulator's path in Cartesian space, we also specify the time corresponding to each path point. 

The change in the end effector pose of the manipulator is determined by the alteration of each joint angle\cite{ref02}. Due to constraints on the angles, velocities, and accelerations of the manipulator's joints, the selection of joint angles at the current time will restrict the range of the manipulator's subsequent poses. In other words, the joint angles at the current sampling time may render the manipulator unable to reach the specified pose at the next sampling time, leading to an interruption. The manipulator used as an example in this study can be controlled through Cartesian pose commands. However, this method only considers the manipulator's current joint positions and does not optimize the joint angle values along the entire path, thus failing to guarantee the complete traversal of the entire path by the manipulator. We will illustrate this in Section IV through experiments. 

Therefore, selecting the appropriate corresponding joint angles for each pose is crucial in the operation of the manipulator. We must consider the entire path as a whole and determine the joint angles corresponding to all sampling points simultaneously.  This involves addressing the inverse kinematics problem of the redundant manipulator, commonly referred to as redundancy resolution.

While calculating the inverse kinematic solutions of a 7-DOF manipulator, the number of unknowns (degrees of freedom of the end effector) is greater than the number of equations (joint angles). Therefore, the inverse kinematics mathematical problem does not have a unique solution, making the inverse kinematics problem complex\cite{ref02}. To ascertain the unique solution to the redundancy resolution problem and ensure the proper functioning of the manipulator, we approach it as an optimization problem with constraints. In doing so, we will introduce a loss function as the objective for optimization.

When specifying the path of the manipulator in Cartesian space, it is not guaranteed that the manipulator can traverse the specified path without breakpoints, because inverse kinematic solutions that comply with joint constraints may not necessarily exist. For such paths, it is possible to divide the path into several subpaths and connect adjacent subpaths through interruption and re-orientation, thus completing the motion along the entire path. In order to ensure the efficiency of the manipulator's motion, we aim to minimize the number of breakpoints and re-orientations during the motion process, considering this as an optimization objective for our algorithm. 

In certain scenarios we consider, such as during ultrasound examinations of the breast, the probe will orient and scan along a circular path\cite{refk}. The selection of the starting point for the circular path does not impact the effectiveness of the ultrasound examination. However, due to constraints on the joint angles of the manipulator, it may affect the frequency of path breakpoints. We aim to optimize the starting position of the circular path concurrently during redundancy resolution calculations, with the goal of minimizing the number of breakpoints and re-orientations required for the path.

We will use a 7-degree-of-freedom manipulator produced by Franka\cite{ref10,ref11} as an example for the algorithm. We obtain an analytical solution for the inverse kinematics of the manipulator by introducing parameters. The redundancy resolution problem is then transformed into a discrete optimization problem to determine the parameters at each sampling point. The constraints on the joint velocity and acceleration of the manipulator are incorporated as constraints in the optimization problem. To solve this optimization problem, dynamic programming is employed. We also consider the scenario where a complete path does not exist. The proposed algorithm can calculate the minimum number of breakpoints required for the path and determine a starting point with the fewest breakpoints for a circular path. Finally, we enhance the efficiency of the algorithm through interpolation and motion compensation.

\subsection{Related Work}

Redundancy resolution can be done in the velocity or position level\cite{ref02}. Many algorithms are designed on the joint velocity level. However, these algotrithms involve online computing the pseudoinverse of the Jacobian matrix associated with the forward kinematics of the controlled manipulator\cite{ref021}, which can only obtain a locally optimal solution.  An analytical methodology of inverse kinematic computation for 7-DOF redundant manipulators with joint limits is proposed by \cite{ref03}. This algorithm obtains all feasible inverse kinematic solutions in the global configuration space. A globally optimal solution to the inverse kinematic problem for a general serial 7-DOF manipulator with revolute joints is proposed in \cite{ref04}, indicating that the kinematic constraints due to rotations can be all generated by the second-degree polynomials. However, the aforementioned references only consider the redundant resolution of individual poses in space, without taking into account the entire path.

A joint parametrization method for redundancy resolution is proposed in \cite{ref12}. In their method, redundant joints are selected appropriately and the joint displacements themselves are regarded as the redundancy parameters, obtaining a closed-form of inverse kinematic solution. This approach works well for the manipulator we use for examination. Other parameterization methods, such as those proposed in \cite{ref03}, can also be selected based on the specific structure of the manipulator.

This paper then uses dynamic programming to determine the parameter values of each point on the path. Dynamic programming is a mathematical method for solving a class of optimization problems. It breaks down complex problems into a series of interconnected and overlapping subproblems, and combines the solutions of these subproblems recursively to obtain the optimal solution to the original problem\cite{ref3,ref4}. In path planning problems, many researchers have used dynamic programming algorithms for optimization. A solution to the problem of minimizing the cost of moving a manipulator along a specified geometric path in the joint space of the manipulator subject to input torque/force constraint is proposed in \cite{ref5}. The algorithm proposed in \cite{ref6} plans minimum energy consumption trajectories of manipulators by iterative dynamic programming. The algorithm proposed in \cite{ref7} divides the entire path into smaller segments, performs interpolation on each segment, and then combines them into a complete path using dynamic programming. By appropriately discretizing the problem, dynamic programming can be employed for path planning of manipulators. A dynamic programming framework for optimal planning of redundant robots along prescribed paths is propsed in \cite{ref020}. In their study, the velocity in Cartesian space is an optimization variable, and thus they constructed constraints on velocity and acceleration in joint space using the Jacobian matrix and Cartesian velocity. In our research, however, the velocity in Cartesian space is fixed, necessitating the exploration of alternative methods to construct constraints.

\cite{refn1} proposed an algorithm using dynamic programming to solve redundancy resolution and developed a ROS implementation based on this approach. However, their algorithm cannot guarantee the smoothness of the path. \cite{refn2} further smoothens the obtained path using an optimal interpolation method after employing dynamic programming to compute redundancy resolution, ensuring that the motion of each joint complies with the constraints. This algorithm requires solving two optimization problems to obtain the final path. Considering the substantial computational and memory requirements of dynamic programming, we also intend to compute redundancy resolution with fewer path points and interpolate the path. Our aim is to employ a simpler interpolation algorithm while ensuring the smoothness of the path and the efficiency of the algorithm.

\cite{refn3} analyzes the conditions for the existence of dynamic programming solutions based on the topological structure of grid graphs, but has not further discussion on how to appropriately partition path when the complete path does not exist.

\subsection{Paper Contribution and Organization}

The main contribution of this paper is the proposal of a globally optimal redundancy resolution algorithm. The algorithm is capable of finding a globally optimal solution for the path that minimizes the cost function. This paper presents three main innovations:

$\bullet$ Through the design of a penalty term, the algorithm is able to find a manipulator path with the least number of breakpoints.

$\bullet$ Specifically for circular paths, an initial point modification algorithm has been designed to further minimize the number of breakpoints during manipulator motion.

$\bullet$ We enhance the efficiency of the algorithm through interpolation while maintaining the smoothness of the path.

This paper is organized as follows: First, the mathematical model of the problem is described in section II. Then, the dynamic programming-based redundancy resolution algorithm is proposed in section III. Finally, we experimentally evaluate the proposed algorithm in section IV.

\section{Preliminaries and problem formulation}

In our work, we will take the Franka 7-DOF manipulator\cite{ref10} for study and experimentation. The host system communicates with the manipulator at a frequency of 1000Hz. The host system is capable of real-time access to various motion parameters of the manipulator, enabling it to send control commands for the motion of the manipulator. Therefore, we choose to discretize the pre-planned path in Cartesian space, ensuring that the sampling interval between adjacent path points is a multiple of the communication period of the manipulator. Fig. 1\cite{ref9} shows the Denavit-Hartenberg parameters of the Franka Emika manipulator, and Table 1 shows the limits of the joints of the manipulator.

\begin{figure}[h]
	\centering
	\includegraphics[width=3.5in]{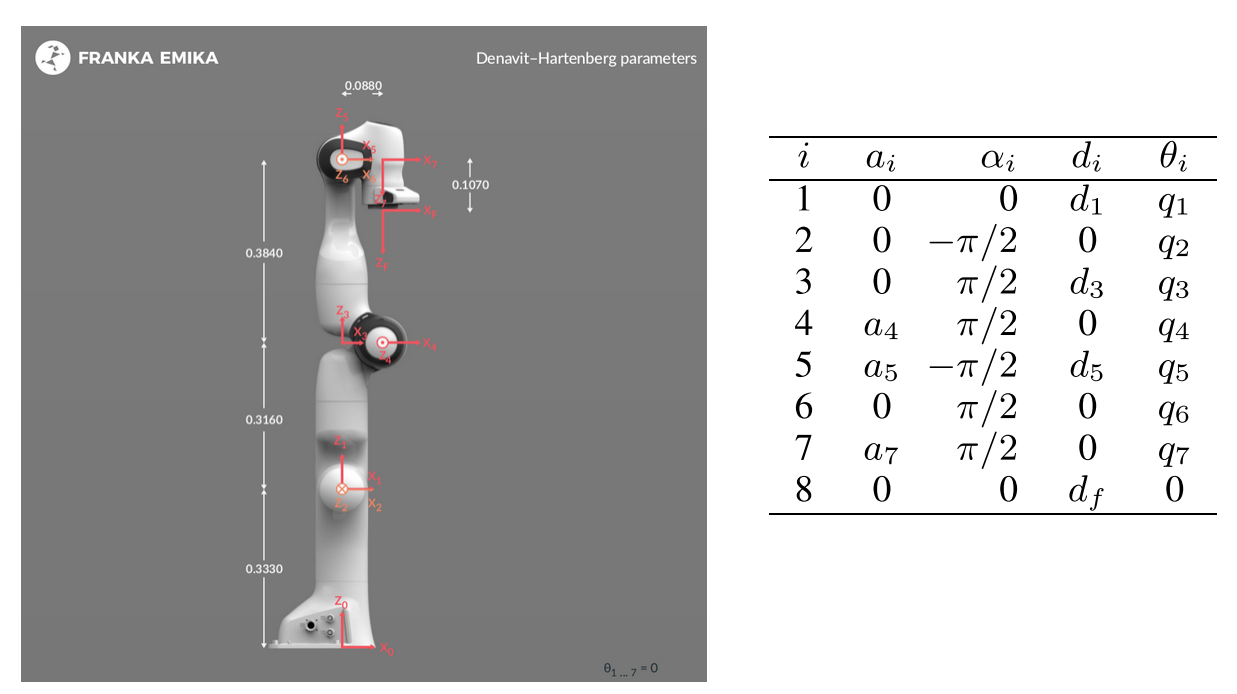}
	\caption{Denavit-Hartenberg frames and table of parameters for the Franka Emika robot. In this figure, joint angles $ q_{1,...,7} = 0 $. The reference frames follow the modified Denavit-Hartenberg convention, $ d_1 = 0.333 \text{m}, d_3 = 0.316 \text{m}, d_5 = 0.384 \text{m}, d_f = 0.107 \text{m}, a_4 = 0.0825 \text{m}, a_5 = -0.0825 \text{m}, a_7 = 0.088\text{m}$.} 
	
	\label{fig_1}
\end{figure}

\begin{table*}
	\caption{Limits of Joints\label{tab:table1}}
	\centering
	\renewcommand{\arraystretch}{1.5}
	\begin{tabular}{c c c c c c c c c}
		\hline
		Name  &  Joint 1 & Joint 2  &  Joint 3  & Joint 4  &  Joint 5  &  Joint 6  &  Joint 7  &  Unit  \\
		\hline
		$ \mathbf{q}_{\max } $ & 2.8973 & 1.7628 & 2.8973 & -0.0698 & 2.8973 & 3.7525 & 2.8973 & rad \\
		$ \mathbf{q}_{\min } $ & -2.8973 & -1.7628 & -2.8973 & -3.0718 & -2.8973 & -0.0175 & -2.8973 & rad \\
		$ \dot{\mathbf{q}}_{\max } $ & 2.1750 & 2.1750 & 2.1750 & 2.1750 & 2.6100 & 2.6100 & 2.6100 & $ \frac{\text{rad}}{\text{s}} $ \\
		$ \ddot{\mathbf{q}}_{\max } $ & 15 & 7.5 & 10 & 12.5 & 15 & 20 & 20 & $ \frac{\text{rad}}{\text{s}^{2}} $ \\
		\hline
	\end{tabular}
\end{table*} 

Firstly, we consider a path of the manipulator in Cartesian space that spans a duration of $ t_{\max} $. For each time $ t $ ranging from 0 to $ t_{\max} $, the end effector of the manipulator has to reach a certain Cartesian pose $ \hat{\mathbf{T}}_{\text{EE}}(t) $, whitch is the $ 4\times 4$ homogeneous transformation matrix of the end effector at the time $ t $. 

During the motion of the manipulator, control commands are sent to the manipulator at a fixed frequency, which means the joint angles of the manipulator can be treated as a discrete function\cite{ref14}. Therefore, we sample the Cartesian path at a fixed time interval $t_0$, which is the communication cycle of the manipulator, and denote the end effector pose of the $ i $th sampling point as $\mathbf{T}_{\text{EE}i}$, where $ i $ represents the index and ranges from 0 to $\frac{t_{\max}}{t_0} $. Denote $\frac{t_{\max}}{t_0}$ as $ n $. By definition, we have: 
\begin{equation}
	\mathbf{T}_{\text{EE}i} = \hat{\mathbf{T}}_{\text{EE}}(t_0\times i),i=0,1,...,n
\end{equation} 

The redundancy resolution to $\mathbf{T}_{\text{EE}i}$ is the joint angle vector $\mathbf{q}_{i}$ of the manipulator that satisfies the equation:
\begin{equation}
	\mathbf{T}_{\text{EE}i} = f(\mathbf{q}_{i})
\end{equation}
where $f$ is the forward kinematics map from the joint angles $\mathbf{q}_{i} $ to the end effector pose $\mathbf{T}_{\text{EE}i}$. Due to the redundancy of the 7-DOF manipulator, the equation above has infinite solutions. In order to ensure the proper functioning of the manipulator, we need to determine a globally optimal solution from the infinite set of inverse kinematic solutions using an algorithm, which entails addressing an optimization problem. To address this optimization problem, it is necessary to define the constraints and the objective function of the optimization problem.

The constraints of the optimization problem lie in the limitations imposed by the hardware of the manipulator on its joint angles, joint velocities, and joint accelerations. The angular volocities and accelerations of joints can be easily calculated by definition:
\begin{equation}
	\begin{aligned}
		\dot{\mathbf{q}}_{i} = \frac{\mathbf{q}_{i}-\mathbf{q}_{i-1}}{t_0}, i = 1,2,...,n
		\\ \ddot{\mathbf{q}}_{i} = \frac{\dot{\mathbf{q}}_{i}-\dot{\mathbf{q}}_{i-1}}{t_0}, i = 2,3,...,n
	\end{aligned}
\end{equation}
where $\dot{\mathbf{q}}_{i}$ is the velocity of the joints at the $ i $th sampling point, and $\ddot{\mathbf{q}}_{i}$ is the acceleration. This allows us to construct the constraints for the optimization problem:
\begin{equation}
	\begin{aligned}
		{q}_{\min,c} \leqslant q_{i,c} \leqslant {q}_{\max,c}, i = 0,1,...,n;c = 1,2,...,7 \\
		|\dot{q}_{i,c}|\leqslant {\dot{q}}_{\max,c}, i = 1,2,...,n;c = 1,2,...,7
		\\ |\ddot{q}_{i,c}|\leqslant {\ddot{q}}_{\max,c} , i = 2,3,...,n;c = 1,2,...,7
	\end{aligned}
\end{equation}
where ${q}_{\min,c}, {q}_{\max,c}, q_{i,c}, \dot{q}_{i,c}, \dot{q}_{\max,c}, \ddot{q}_{i,c}, \ddot{q}_{\max,c} $ represent the $c$-th components of vectors $\mathbf{q}_{\min},\mathbf{q}_{\max}, \mathbf{q}_i, \dot{\mathbf{q}}_i, \dot{\mathbf{q}}_{\max}, \ddot{\mathbf{q}}_i, \ddot{\mathbf{q}}_{\max} $ respectively.

In terms of the loss function, we use $ \{\mathbf{q}_a\} $ to represent the subsequence $ \mathbf{q}_0 $, $ \mathbf{q}_1 $,..., $ \mathbf{q}_a $\ of length $a$, where $ a $ is the number of sampling points in the sequence, so the redundancy resolution of the entire path is represented as $ \{\mathbf{q}_n\} $. In the path, there may be discontinuity points where the manipulator does not move directly from the current joint angle to the next one, but instead halts its current motion before proceeding to the next joint angle to initiate a new movement. Let $cont_i$ represent whether the manipulator’s motion is interrupted when moving from $q_{i-1}$ to $q_i$. If the motion is interrupted, $cont_i$ equals 1;otherwise, $cont_i$ equals 0. We define the loss function as:
\begin{equation}
	L(\{\mathbf{q}_a\}) =  \sum_{i=1}^a{cont_i||\mathbf{q}_{i}-\mathbf{q}_{i-1}||_2^2+(1-cont_i) M} 
\end{equation}
where $ \{\mathbf{q}_a\} $ is the sequence $ \mathbf{q}_0 $, $ \mathbf{q}_1 $,..., $ \mathbf{q}_a $ with $ a $ sampling points and $M$ is a large constant that satisfies $M > n||\mathbf{q}_{\max}-\mathbf{q}_{\min}||^2_2$. The selection of $M$ in this manner is due to the fact that dividing $L(\{\mathbf{q}_a\})$ by $M$ yields the number of discontinuities required for $ \{\mathbf{q}_a\} $. 

$ L(\{\mathbf{q}_n\}) $ is the loss function of the entire path. Since the time interval between adjacent sampling points is fixed, sum of squared differences in joint angles can measure the square sum of angular velocities. A smaller angular velocity can reduce the energy consumption and mechanical wear of the manipulator, while also improving its safety. By introducing the penalty term $M$, the algorithm is able to obtain a path with minimal breakpoints. By minimizing the loss function $ L(\{\mathbf{q}_n\}) $, the algorithm can provide a globally optimal solution to the optimization problem, serving as the final motion path for the manipulator. 

Moreover, the chosen loss function is computationally straightforward and well-suited for demonstrating the algorithm in this paper. Depending on the specific requirements of practical applications, alternative loss functions can be employed. 

\section{Redundancy Resolution for Path Planning}

In this section, we will use dynamic programming to obtain the redundancy resolution $\{\mathbf{q}_n\}$ for the given Cartesian poses $\{\mathbf{T}_{\text{EE}n}\}$. 

\subsection{Inverse Kinematic Solution of Redundant Manipulators}

Before seeking the redundancy resolution for the entire path, it is worthwhile to first consider the inverse kinematic solution corresponding to a single pose $\mathbf{T}_\text{EE}$ in Cartesian space. Due to the redundancy of the redundant manipulator, there exist infinitely many inverse kinematic solutions. By employing the parameterization technique, we can derive closed-form inverse kinematics solutions for the manipulator. According to  \cite{ref12}, a certain joint angle can be fixed as a parameter. The algorithm proposed in \cite{ref15} calculates the inverse kinematic solutions of the Franka robot by taking $q_7$, the joint angle of the 7th joint of the manipulator, as a parameter. When a solution exists, there are three possible scenarios that lead to multiple solutions, resulting in up to 8 sets of solutions. These multiple solutions arising from the three scenarios can be mutually converted when they exist. By considering the constraints on each joint angle, an analysis is conducted to exclude the multiple solutions while preserving the existence of the solutions through the imposition of restrictions on the joint angles. From this, we can obtain a bijective inverse kinematic mapping $f^{-1}_{q_7}$ within the workspace of the 6-DOF manipulator:
\begin{equation}
	[q_1, q_2, ..., q_6]^T = f^{-1}_{q_7}(\mathbf{T}_\text{EE})
\end{equation}
Thus, we have obtained the parameterized inverse kinematic solution for the 7-DOF manipulator:
\begin{equation}
	\begin{aligned}
		\mathbf{q} & = \tilde{f}^{-1}(\mathbf{T}_\text{EE}, q_7)\\
		& = \begin{bmatrix}
			f^{-1}_{q_7}(\mathbf{T}_\text{EE})	\\
			q_7
		\end{bmatrix}
	\end{aligned}	
\end{equation}

For other manipulators with higher degrees of freedom, similar parameter selection methods can be employed, albeit with a corresponding increase in the dimensionality of the parameters.

In this paper, we will adopt another discretization approach: dividing the range of $q_7$ into $m-1$ equal parts, thereby restricting the values of $q_7$ to $m$ discrete values $\{a_1, a_2, \ldots, a_m\}$. By discretizing, we limit the final path of the manipulator to a finite set of $m^{(n+1)}$ possible configurations. For manipulators with higher than 7 degrees of freedom, discretizing the values of each parameter similarly allows the problem to be transformed into one analogous to a 7-degree-of-freedom manipulator. The parameter values $a_i$ at this point will be replaced by the vector $\mathbf{a}_i$. Clearly, as $m$ increases, we consider a larger number of possibilities, leading to improved algorithm performance. When $m$ is small, the resulting loss function of the obtained solution will be larger, but it still presents a viable path for the normal operation of the manipulator, due to its compliance with all the constraints imposed on the joints. 

For given values of $\mathbf{T}_{\text{EE}i}$ and $a_j$, the inverse kinematic solution not necessarily exist due to various restrictions on the manipulator joint angles and mechanical structure. The following test path serves as an example:

\begin{equation}
	\hat{\mathbf{T}}_{\text{EE1}}(t)  = \begin{bmatrix}
		\cos(\theta(t))	& \sin(\theta(t)) & 0 &0.6+0.1\cos(\theta(t)) \\
		\sin(\theta(t))	& -\cos(\theta(t)) & 0 & 0.1\sin(\theta(t))\\
		0	& 0 & -1 & 0.1\\
		0	& 0 &0  &1
	\end{bmatrix}
\end{equation}
where $\theta(t) = \frac{2\pi}{ t_{\max}  }t - \sin(\frac{2\pi}{ t_{\max}  }t ) - \pi, 0 \leqslant t \leqslant t_{\max}$ and the duration $t_{\max}$ is 10 seconds. 

The existence of solutions corresponding to each $ \hat{\mathbf{T}}_{\text{EE1}}(t) $ and $q_7$ is depicted in a figure, where the blue regions indicate the existence of the inverse kinematic solutions, as shown in Figure 2. Denote $\tilde{f}^{-1}(\mathbf{T}_{\text{EE}i}, a_j) $ as $\bar{\mathbf{q}}_{i,j}$ if exist.

\begin{figure}[h]
	\centering
	\includegraphics[width=3.2in]{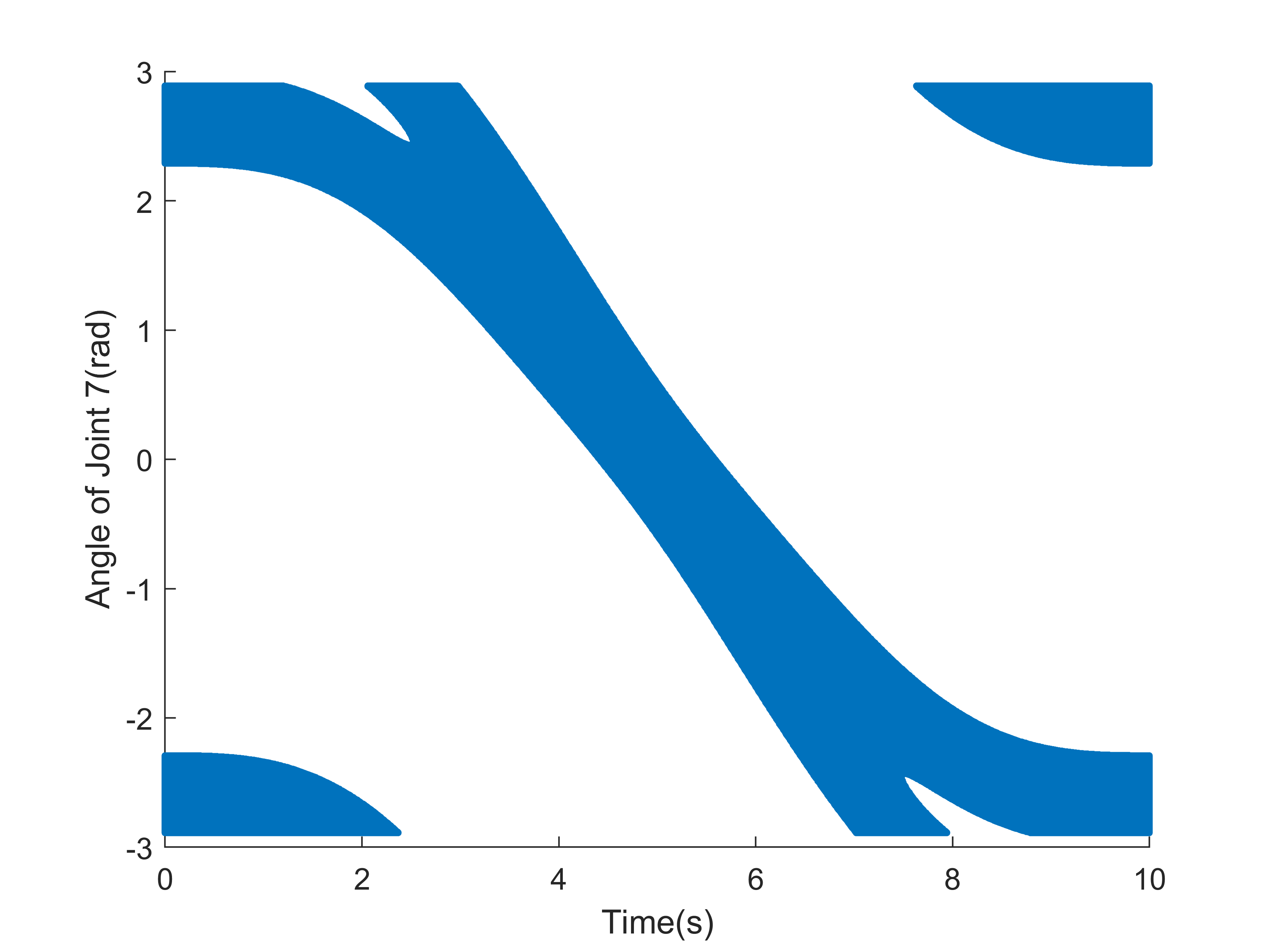}
	\caption{The existence of solutions corresponding to each $ \hat{\mathbf{T}}_{\text{EE1}}(t) $ and $q_7$ is depicted in a figure, where the blue regions indicate the existence of the corresponding inverse kinematic solutions $\tilde{f}^{-1}(\hat{\mathbf{T}}_{\text{EE1}}(t), q_7) $.} 
	
	\label{fig_2}
\end{figure}

Taking into account the constraints on joint velocity and acceleration, the curve of $q_7$ over time will be a continuous curve within the blue portion of the figure of the existence of inverse kinematic solutions. For certain paths, such as another test path:

\begin{equation}
	\hat{\mathbf{T}}_{\text{EE2}}(t)  = \begin{bmatrix}
		\cos(\theta(t))	& \sin(\theta(t)) & 0 &0.6+0.1\cos(\theta(t)) \\
		\sin(\theta(t))	& -\cos(\theta(t)) & 0 & 0.1\sin(\theta(t))\\
		0	& 0 & -1 & 0.1\\
		0	& 0 &0  &1
	\end{bmatrix}
\end{equation}
where $\theta(t) = \frac{2\pi}{ t_{\max}  }t, 0 \leqslant t \leqslant t_{\max}$ and the duration $t_{\max}$ is 10 seconds. The existence of solutions corresponding to each $ \hat{\mathbf{T}}_{\text{EE2}}(t) $ and $q_7$ is depicted in Figure 3:

\begin{figure}[h]
	\centering
	\includegraphics[width=3.2in]{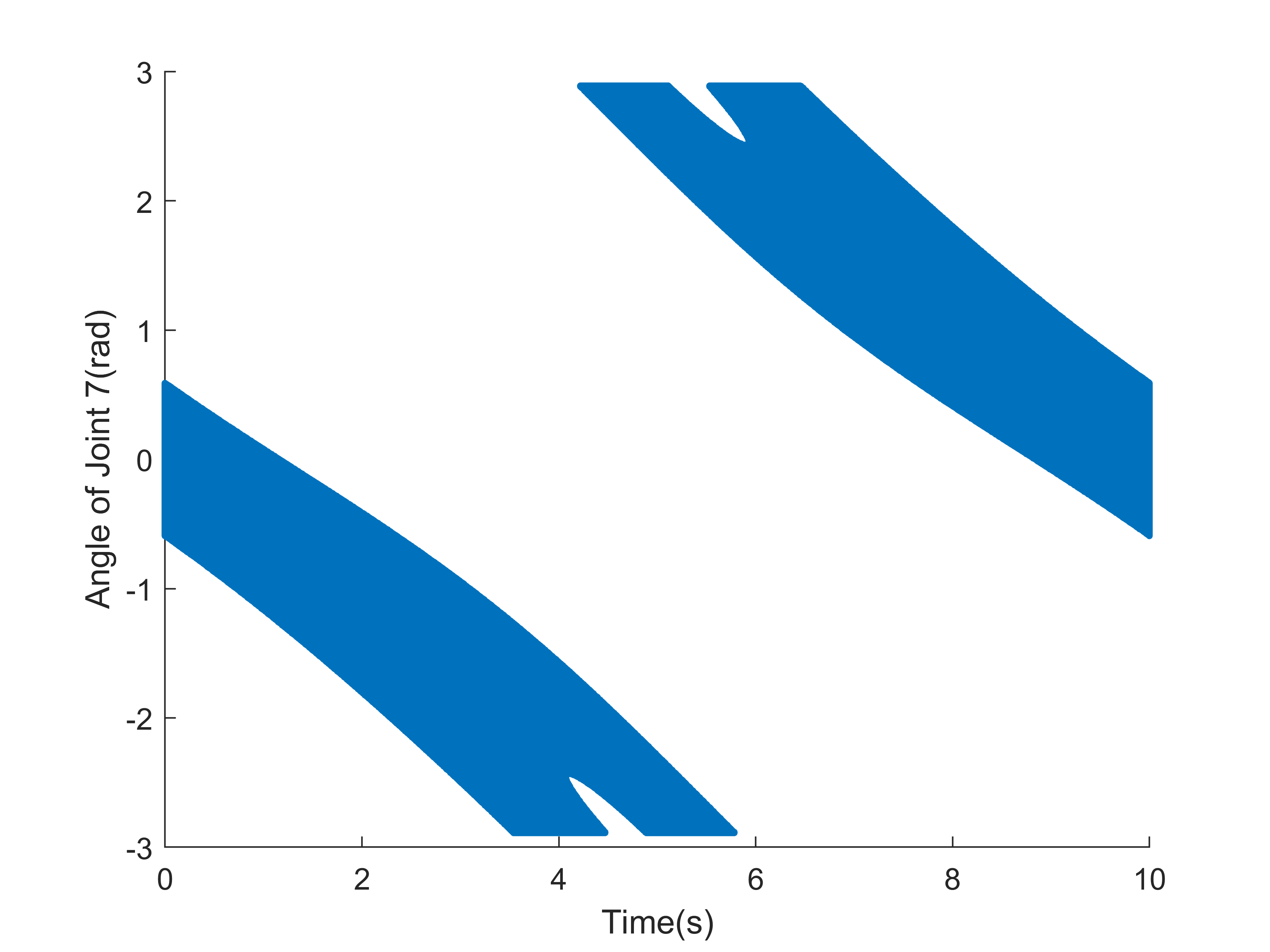}
	\caption{The existence of solutions corresponding to each $ \hat{\mathbf{T}}_{\text{EE2}}(t) $ and $q_7$ is depicted in a figure, where the blue regions indicate the existence of the corresponding inverse kinematic solutions $\tilde{f}^{-1}(\hat{\mathbf{T}}_{\text{EE2}}(t), q_7) $.} 
	
	\label{fig_3}
\end{figure} 

A continuous curve of $q_7$ does not exist, therefore the manipulator cannot traverse from the start to the end along the path without interruption. From this example, it is evident that in designing the algorithm and the loss function, we should also consider the possibility of path interruption and minimize the number of breakpoints. We can intuitively determine the number of breakpoints required to traverse a given path through the topological structure of the graph. But we prefer to directly compute the minimal number of breakpoints using the dynamic programming algorithm without additional analysis. We also aim to determine the optimal selection of interruption points to minimize the global cost function, rather than simply interrupting when the manipulator is unable to continue operating. The design of our cost function $L(\{\mathbf{q}_{i}\})$ aims to prioritize minimizing the number of breakpoints while simultaneously reducing the remaining cost terms as much as possible.

\subsection{Dynamic Programming-Based Redundancy Resolution}

Let $ \tilde{L}{(i,j,k)} $ represent the minimum loss function of $\{\mathbf{q}_{i}\}$ that reaches $ \mathbf{q}_{(i-1,k)} $ at the ${(i-1)}$th sampling point and then stops at $ \mathbf{q}_{(i,j)} $ at the $i$-th sampling point. We have: 
\begin{equation}
	\tilde{L}{(i,j,k)} = \min \{L(\{\mathbf{q}_{i}\}) | \mathbf{q}_{i} = \bar{\mathbf{q}}_{i,j},\mathbf{q}_{i-1} = \bar{\mathbf{q}}_{i-1,k} \}
\end{equation}
where $1\leqslant i\leqslant n$.  

Denote the $c$-th component of $ \bar{\mathbf{q}}_{i,j} $ as $\bar{q}_{i,j,c} $. When the velocity constraints are not satisfied, which means there is a $c$ that satisfies:
\begin{equation}
	|\dot{q}_{i,c}|=|\frac{\bar{q}_{i,j,c}-\bar{q}_{i-1,k,c}}{t_0}|> {\dot{q}}_{\max,c}
\end{equation}
$\tilde{L}{(i,j,k)}$ is set to infinite. Otherwise, we have:
\begin{equation}
	\tilde{L}(1,j,k)  =  ||\bar{\mathbf{q}}_{1,j}-\bar{\mathbf{q}}_{0,k}||_2^2
\end{equation}

To iteratively compute $\tilde{L}{(i,j,k)}$ when constraints are satisfied and $i\geqslant 2$, we will consider the minimum loss function $ \hat{L}{(i,j,k,p)} $ of $\{\mathbf{q}_{i}\}$ that passes through $ \bar{\mathbf{q}}_{i-2,p} $, $  \bar{\mathbf{q}}_{i-1,k} $, and stops at $ \bar{\mathbf{q}}_{i,j} $ at the $i$-th sampling point: 
\begin{equation}
	\hat{L}{(i,j,k,p)} = \min\{L\{\mathbf{q}_{i}\} | \mathbf{q}_{i} = \bar{\mathbf{q}}_{i,j},\mathbf{q}_{i-1} =  
	\bar{\mathbf{q}}_{i-1,k},\mathbf{q}_{i-2} = \bar{\mathbf{q}}_{i-2,p}\}		
\end{equation}
where $2\leqslant i\leqslant n$. 

When the acceleration and velocity constraints are satisfied, which means:
\begin{equation}
	\begin{aligned}
		|\dot{q}_{i-1,c}|=|\frac{\bar{q}_{i-1,k,c}-\bar{q}_{i-2,p,c}}{t_0}|\leqslant {\dot{q}}_{\max,c},  c = 1,2,...,7
		\\
		|\dot{q}_{i,c}|=|\frac{\bar{q}_{i,j,c}-\bar{q}_{i-1,k,c}}{t_0}|\leqslant {\dot{q}}_{\max,c}, c = 1,2,...,7
		\\ |\ddot{q}_{i,c}|=|\frac{\dot{q}_{i,c}-\dot{q}_{i-1,c}}{t_0}|\leqslant {\ddot{q}}_{\max,c} ,  c = 1,2,...,7
	\end{aligned}
\end{equation}
we have:
\begin{equation}
	\begin{aligned}
		\hat{L}{(i,j,k,p)}  = & \tilde{L}{(i-1, k, p)} + ||\bar{\mathbf{q}}_{i,j}-\bar{\mathbf{q}}_{i-1,k}||_2^2
	\end{aligned}	
\end{equation}
Otherwise, the manipulator cannot pass through $ \bar{\mathbf{q}}_{i-2,p} $, $  \bar{\mathbf{q}}_{i-1,k} $, and stops at $ \bar{\mathbf{q}}_{i,j} $ the path without interruption. To avoid redundant considerations, we stipulate the path breaks between  $ \bar{\mathbf{q}}_{i-2,p} $ and $  \bar{\mathbf{q}}_{i-1,k} $. In this scenario, we have:	
\begin{equation}
	\hat{L}(i,j,k,p) = M + \min_{r=1}^m \tilde{L}(i-2,p,r) +||\bar{\mathbf{q}}_{i,j}-\bar{\mathbf{q}}_{i-1,k}||_2^2
\end{equation}	
Simultaneously, for the generation of the path in the subsequent algorithm, we supplement the definition:
\begin{equation}
	\tilde{L}(i-1,k,p) = M + \min_{r=1}^m\tilde{L}(i-2,p,r)
\end{equation} 
\begin{equation}  
	\hat{L}(i-1,k,p,r) = \left\{\begin{matrix}  
		{M + \tilde{L}(i-2,p,r), } { r=\arg\min_{r=1}^m\tilde{L}(i-2,p,r)}\\  
		{+\infty ,} \text{otherwise}  
	\end{matrix}\right.	  
\end{equation}  

$\bar{\mathbf{q}}_{i-2,p}$ will serve as the endpoint of the manipulator's original motion path, after which the manipulator moves to $\bar{\mathbf{q}}_{i,j}$ to commence the movement along a new path.

And obviously we have:	
\begin{equation}
	\tilde{L}{(i,j,k)} = \min_{p=1}^m \hat{L}{(i,j,k,p)}
\end{equation}	
where $2\leqslant i\leqslant n$. 

Iteratively, for each $ {(i,j,k)} $, we calculate the minimum value $ \tilde{L}{(i,j,k)} $  and proceed to the next recursion. Then, we can find the path with the minimum loss function by traversing $\tilde{L}(n,j,k)$. By seeking the $p$ that minimizes $\hat{L}{(i,j,k,p)}$, we can obtain the previous joint angles iteratively. Algorithm 1 shows the solution.

\begin{algorithm}
	\caption{Dynamic Programming-Based Redundancy Resolution}\label{alg:alg1}
	\begin{algorithmic}
		\STATE 
		\STATE {\text{Set }}$i = 1$
		\STATE {\text{For}} $1\leqslant j,k\leqslant m,\text{ calculate } \tilde{L}(1,j,k)$
		
	\STATE 
	\STATE {\text{For }}$(1<i\leqslant n )$
	\STATE \hspace{0.5cm}\text{For} $1\leqslant j,k\leqslant m$
	\STATE	\hspace{1cm}\text{if} $\left| \bar{q}_{i,j,c}-\bar{q}_{i-1,k,c} \right|
		\leqslant {\dot{q}_{\max ,c}} \times t_0, \forall c$
	\STATE \hspace{1.5cm}{\text{For}} $1\leqslant p\leqslant m$ 
	\STATE \hspace{2cm}{\text{Calculate}} $ \hat{L}{(i,j,k,p)} $
	\STATE \hspace{1.5cm}$ \tilde{L}{(i,j,k)} = \min_{p=1}^m \hat{L}{(i,j,k,p)} $
		
	\STATE \hspace{1cm}\text{else }$\tilde{L}{(i,j,k)} = +\infty$
	\STATE
	\STATE {\text{Find }}$j_n,j_{n-1} $\text{, such that: }
	\STATE \hspace{0.5cm}$\tilde{L}(n,j_n,j_{n-1}) = \min_{j,k=1}^m \tilde{L}(n,j,k)$
	\STATE {\text{If }}$  \min_{j,k=1}^m \tilde{L}(n,j,k) = +\infty$
	\STATE \hspace{0.5cm}{\text{There is no solution. }} 
	\STATE {\text{Set }}$\mathbf{q}_n = \bar{\mathbf{q}}_{n,j_n},\mathbf{q}_{n-1} = \bar{\mathbf{q}}_{n-1,j_{n-1}}$
	\STATE {\text{For }}$(n-1>i\geqslant 0)$
	\STATE \hspace{0.5cm}$j_i = \text{argmin}_{p=1}^m \hat{L}(i+2,j_{i+2},j_{i+1},p)$
	\STATE \hspace{0.5cm}{\text{Set }}$\mathbf{q}_{i} = \bar{\mathbf{q}}_{i,j_i}$
	\end{algorithmic}
	\label{alg1}
\end{algorithm}
The calculated $ \{\mathbf{q}_n\} $ is the globally optimal redundancy resolution with its loss function $ L(\{\mathbf{q}_n\}) = \min_{j,k=1}^m \tilde{L}(n,j,k)$.

\subsection{Starting Point Modification}

For circular paths with identical starting and ending poses, if their redundancy resolution has breakpoints, we can reduce the number of breakpoints by selecting another point on the path as the new starting point.

Assuming that a circular path $\{\mathbf{T}_{\text{EE}n}\}$ necessitates at least $z$ discontinuities, we can alter the starting point to $\mathbf{T}_{\text{EE}a}$, potentially reducing the number of discontinuities in the path by one. Let the modified starting point require $z'$ discontinuities, corresponding to the path $\{^a\mathbf{q}_n\}$. It is evident that $^a\mathbf{q}_i$ corresponds to the manipulator's pose $\mathbf{T}_{\text{EE}(a+i)\%n}$. Initiating motion from $^a\mathbf{q}_{n-a}$ to $^a\mathbf{q}_n$, then interrupting the motion, adjusting the joint angles to $^a\mathbf{q}_1$, and subsequently moving to $^a\mathbf{q}_{n-a}$ results in the termination of motion. The corresponding manipulator end-effector path remains as $\{\mathbf{T}_{\text{EE}n}\}$. The sum of the discontinuities in the two segments of the manipulator's motion corresponds to the number of discontinuities in $\{^a\mathbf{q}_n\}$, which is $z'$. If $^a\mathbf{q}_n$ and $^a\mathbf{q}_0$ are different, the motion interruption between $^a\mathbf{q}_n$ and $^a\mathbf{q}_1$ introduces a new interruption. Therefore, the number of discontinuities in this new path is either $z'+1$ or $z'$. Since the circular path $\{\mathbf{T}_{\text{EE}n}\}$ necessitates at least $z$ discontinuities, we have $z'$ $\geq$ $z$ - 1. Consequently, altering the starting point reduces the number of discontinuities by at most one.

When $z'$ = $z - 1$, reverting the starting point of $\{^a\mathbf{q}_n\}$ back to $\mathbf{T}_{\text{EE}0}$ as described earlier results in a joint space path corresponding to $\{\mathbf{T}_{\text{EE}n}\}$ with $z$ breakpoints, where $^a\mathbf{q}_n$ to $^a\mathbf{q}_1$ constitutes one of these breakpoints. Thus, to minimize the number of breakpoints, we should consider the new starting point after the interruption in the path corresponding to $\{\mathbf{T}_{\text{EE}n}\}$ as $\mathbf{T}_{\text{EE}a}$.

The path $\{^a\mathbf{q}_n\}$ is segmented by $z'$ discontinuities, resulting in $z'+1$ continuous segments. Choosing any of these segments as the new starting point yields a new path with $z'$ discontinuities. It is reasonable to stipulate that we take the new starting point after the first interruption in $\{\mathbf{T}_{\text{EE}n}\}$ as the starting point for the path, as selecting a starting point after subsequent breakpoints will not affect the total number of breakpoints in the new path.

Next, we will introduce a minor modification to Algorithm 1 to enable the adjustment of the starting point. Initially, we duplicate the original path, defining $\mathbf{T}_{\text{EE}n+i}=\mathbf{T}_{\text{EE}i}$ for all $i$. We then extract a continuous sequence of $n+1$ points from $\{\mathbf{T}_{\text{EE}2n}\}$, representing the new path after the starting point modification. Subsequently, Algorithm 1 is applied to $\{\mathbf{T}_{\text{EE}2n}\}$, and we adjust Equation (17) as follows:	
\begin{equation}
	\hat{L}(i,j,k,p) = \left\{\begin{matrix}
		n M + \tilde{L}+(i-1)M,  \tilde{L}<n M\\
		n M + \tilde{L},otherwise
	\end{matrix}\right.
\end{equation} 
where $\tilde{L} = \min_{r=1}^m \tilde{L}(i-2,p,r) $

The modified cost function can not only compute the minimum number of breakpoints required for the manipulator to pass through $ \bar{\mathbf{q}}_{i-2,p} $, $  \bar{\mathbf{q}}_{i-1,k} $, and stop at $ \bar{\mathbf{q}}_{i,j} $ via dividing it by $n\cdot M$, but also determine the minimum index of the starting point after the first interruption given the condition of minimizing the number of breakpoints via dividing the remainder by $M$. This implies the computation of the maximum number of path points the manipulator traverses starting from the newly adjusted path's starting point to reach $ \bar{\mathbf{q}}_{i,j} $.

After computing $\tilde{L}{(i,j,k)}$ using equation (20), we can determine the minimum index of the starting point after the first interruption and denote it as $z{(i,j,k)}$. For $n+1 \leqslant i \leqslant2n$, and for all $i, j$, if $z_{(i,j,k)}+n=i$ and $\tilde{L}_{(i,j,k)}<n(z+1)M$, then there are a total of $n+1$ points from $\mathbf{T}_{\text{EE}z_{(i,j,k)}}$ to $\mathbf{T}_{\text{EE}i}$, with only $z-1$ breakpoints in between. Therefore, designating $\mathbf{T}_{\text{EE}z_{(i,j,k)}}$ as the new starting point of the path can reduce the number of breakpoints required by 1. Algorithm 2 shows the solution.

\begin{algorithm}
	\caption{Dynamic Programming-Based Starting Point Modification}\label{alg:alg2}
	\begin{algorithmic}
		\STATE 
	\STATE {\text{Set }}$i = 1$
		\STATE {\text{For}} $1\leqslant j,k\leqslant m,\text{ calculate } \tilde{L}(1,j,k)$
		
	\STATE 
	\STATE {\text{For }}$(1<i\leqslant 2n )$
		\STATE \hspace{0.5cm}\text{For} $1\leqslant j,k\leqslant m$
		\STATE	\hspace{1cm}\text{if} $\left| \bar{q}_{i,j,c}-\bar{q}_{i-1,k,c} \right|
		\leqslant {\dot{q}_{\max ,c}} \times t_0, \forall c$
		\STATE \hspace{1.5cm}{\text{For}} $1\leqslant p\leqslant m$ 
		\STATE \hspace{2cm}{\text{Calculate}} $ \hat{L}{(i,j,k,p)} $
	\STATE \hspace{1.5cm}$ \tilde{L}{(i,j,k)} = \min_{p=1}^m \hat{L}{(i,j,k,p)} $
		
		\STATE \hspace{1cm}\text{else }$\tilde{L}{(i,j,k)} = +\infty$
	\STATE
	\STATE {\text{For }}$(n+1\leqslant  i\leqslant 2n )$
		\STATE \hspace{0.5cm}\text{For} $1\leqslant j,k\leqslant m$
		\STATE	\hspace{1cm}\text{Calculate}$z_{(i,j,k)}$ \text{and}
		\STATE	\hspace{1cm}\text{if }$z_{(i,j,k)}+n=i$ \text{and} $\tilde{L}_{(i,j,k)}<n(z+1)M$ 
		\STATE \hspace{1.5cm}$\mathbf{T}_{\text{EE}z_{(i,j,k)}}${can be a new starting point} 
	\end{algorithmic}
	\label{alg2}
\end{algorithm}

\subsection{Interpolation}

In practical applications, taking the Franka manipulator as an example, the communication frequency between the manipulator and the host is 1000Hz. This implies that if we were to rigorously calculate the joint angles of the manipulator at each communication instance, we would need to consider 1000 path points per second, significantly increasing both memory and computational burdens. Therefore, in order to enhance the efficiency of the algorithm, we aim to reduce the number of path points for redundancy resolution computation and employ interpolation to determine the joint angles corresponding to each communication point. 

If linear interpolation is employed to interpolate the joint angles of the intermediate sampling points, according to the algorithm, the angle and velocity constraints of each joint will be satisfied, but the acceleration constraints will not be met. This leads to the joints being unable to reach the angles calculated by the interpolation, resulting in errors. Simultaneously, the manipulator may exhibit certain mechanical inaccuracies during operation, leading to discrepancies between the actual joint angles and the desired values, thereby affecting the constraints on velocity or acceleration. We employ a real-time interpolation algorithm to compensate for these errors and reduce the discrepancy between the actual path and the desired path. During communication between the host system and the manipulator, real-time readings of the current joint angles $\mathbf{q}_{now}$, velocities $\dot{\mathbf{q}}_\text{now}$, and accelerations $\ddot{\mathbf{q}}_\text{now}$ of joints are obtained. The expected joint angles $\mathbf{q}_{i+1}$ at the next sampling point are calculated using the proposed algorithm before the motion of the manipulator. Subsequently, the remaining time before the next sampling point, denoted as $t_\text{r}$, is used to compute the expected velocity $\dot{\mathbf{q}}_\text{d}$, acceleration $\ddot{\mathbf{q}}_\text{d}$ and jerk $\dddot{\mathbf{q}}_\text{d}$ for the next communication cycle:

\begin{equation}
	\begin{aligned}
		\dot{\mathbf{q}}_\text{d}& = \frac{\mathbf{q}_{i+1}-\mathbf{q}_\text{now}}{t_\text{r}}\\
		\ddot{\mathbf{q}}_\text{d}& = \frac{\dot{\mathbf{q}}_\text{d}-\dot{\mathbf{q}}_\text{now}}{t_0}\\
		\dddot{\mathbf{q}}_\text{d}& = \frac{\ddot{\mathbf{q}}_\text{d}-\ddot{\mathbf{q}}_\text{now}}{t_0}
	\end{aligned}	
\end{equation}
where $t_0$ is the communication period of the manipulator.

Subsequently, we sequentially examine whether the jerk, acceleration, and velocity exceed their respective limits. If any of these values surpass the limits, they need to be adjusted to their corresponding maximum values. Subsequently, the other two motion parameters need to be recalculated until all motion parameters comply with the constraints. 

Considering the constraint on jerk, the manipulator may not be able to change the acceleration direction in a short time. After accelerating, if the manipulator needs to decelerate in a certain joint, there will still be a period of acceleration. This acceleration period might cause the joint angle to exceed its velocity limits. Therefore, a "cautionary value" needs to be set for the joint velocities. When the velocity surpasses this limit, the manipulator must reduce its acceleration and begin deceleration. 

Similarly, due to the constraint on manipulator acceleration, when the manipulator velocity needs to change direction, there will still be a short deceleration along the current direction. This deceleration may cause the joint angles to exceed their limits. Consequently, when using the algorithm to solve the inverse kinematics, the joint angle limits need to be reduced to accommodate space for deceleration.

In addition, we also need to consider how to smoothly bring the manipulator to a stop as the motion approaches its end. This requires the manipulator's joint velocity, acceleration, and jerk to be zero as the motion stops. In order to smoothly bring the manipulator's velocity and acceleration to zero, additional constraints need to be satisfied:
\begin{equation}
	\begin{aligned}
		&\ddot{\mathbf{q}}\leqslant n_0 t_0\dddot{\mathbf{q}}_\text{max} \\
		&\dot{\mathbf{q}}\leqslant n_0 t_0\ddot{\mathbf{q}}_\text{max} -\frac{\ddot{\mathbf{q}}_\text{max}^2}{2 \dddot{\mathbf{q}}_\text{max}}
	\end{aligned}
\end{equation}
where $n_0$ is the number of remaining communication cycles.

Thus, we have constructed the motion compensation algorithm for manipulator control. At each communication cycle, the final manipulator control algorithm is as described in Algorithm 3:

\begin{algorithm}
	\caption{Interpolation Algorithm}\label{alg:alg3}
	\begin{algorithmic}
		\STATE 
		\STATE {\text{Read the elapsed time $t$ of the motion}}
		
	\STATE {\text{Calculate the remaining time $t_\text{r}$ before ${\mathbf{T}}_{\text{EE}i+1}$}}
		\STATE {$\ddot{\mathbf{q}}_\text{max'} = \max(\ddot{\mathbf{q}}_\text{max}, n_0 t_0\dddot{\mathbf{q}}_\text{max} )$}
		\STATE {$\dot{\mathbf{q}}_\text{max'} = \max(\dot{\mathbf{q}}_\text{max}, n_0 t_0\ddot{\mathbf{q}}_\text{max} -\frac{\ddot{\mathbf{q}}_\text{max}^2}{2 \dddot{\mathbf{q}}_\text{max}})$}
		\STATE {\text{Read $\mathbf{q}_\text{now}$,  $\dot{\mathbf{q}}_\text{now}$, and $\ddot{\mathbf{q}}_\text{now}$}}
		\STATE {$\dot{\mathbf{q}}_\text{d} = \frac{\mathbf{q}_{i+1}-\mathbf{q}_\text{now}}{t_\text{r}}$}
		\STATE {$\ddot{\mathbf{q}}_\text{d} = \frac{\dot{\mathbf{q}}_\text{d}-\dot{\mathbf{q}}_\text{now}}{t_0}$}
		\STATE {$\dddot{\mathbf{q}}_\text{d} = \frac{\ddot{\mathbf{q}}_\text{d}-\ddot{\mathbf{q}}_\text{now}}{t_0}$}
		\STATE {*For $1\leqslant c\leqslant 7$}
		\STATE \hspace{0.5cm} {If ${\dddot{q}}_{\text{d}c}> {\dddot{q}}_{\text{max}c}$ }
		\STATE \hspace{0.93cm} {${\dddot{q}}_{\text{d}c}= {\dddot{q}}_{\text{max}c}$, ${\ddot{q}}_{\text{d}c}={\ddot{q}}_{\text{now}c}+{\dddot{q}}_{\text{d}c}\cdot t_0,$ }
		\STATE \hspace{1cm} {${\dot{q}}_{\text{d}c}={\dot{q}}_{\text{now}c}+{\ddot{q}}_{\text{d}c}\cdot t_0$, return to *}
		\STATE \hspace{0.5cm} {If ${\ddot{q}}_{\text{d}c}> {\ddot{q}}_{\text{max'}c}$ }
		\STATE \hspace{1cm} {${\ddot{q}}_{\text{d}c}= {\ddot{q}}_{\text{max'}c}$, ${\dddot{q}}_{\text{d}c}=\frac{{\ddot{q}}_{\text{now}c}-{\ddot{q}}_{\text{d}c}}{t_0},$ }
		\STATE \hspace{1cm} {${\dot{q}}_{\text{d}c}={\dot{q}}_{\text{now}c}+{\ddot{q}}_{\text{d}c}\cdot t_0$, return to *}
		\STATE \hspace{0.5cm} {If ${\dot{q}}_{\text{d}c}> {\dot{q}}_{\text{max'}c}$}
		\STATE \hspace{1cm} {${\dot{q}}_{\text{d}c}= {\dot{q}}_{\text{max'}c}$, ${\ddot{q}}_{\text{d}c}=\frac{{\dot{q}}_{\text{now}c}-{\dot{q}}_{\text{d}c}}{t_0},$ }
		\STATE \hspace{0.93cm} {${\dddot{q}}_{\text{d}c}=\frac{{\ddot{q}}_{\text{now}c}-{\ddot{q}}_{\text{d}c}}{t_0}$, return to *}
		\STATE {$\mathbf{q}_\text{d} = \mathbf{q}_\text{now} + t_0\dot{\mathbf{q}} $}
		\STATE {Return $\mathbf{q}_\text{d}$ as the expected joint angle for the next communication point.}
		
	\end{algorithmic}
	\label{alg3}
\end{algorithm}

\section{Results}

To verify the feasibility of the proposed algorithm, we selected the Franka Emika manipulator for the research and testing of our algorithm. 

\subsection{Comparison with the Franka Cartesian pose generator}

The manipulator has a Cartesian pose generator that can calculate the inverse kinematic solution of the Cartesian pose based on the current state of the manipulator and control its movements accordingly at a fixed frequency. This algorithm can serve as an example of a local inverse kinematics solution algorithm, demonstrating the limitations of such algorithms and highlighting the advantages of employing dynamic programming to consider the entire path. We utilize the Cartesian pose generator and the proposed algorithm to calculate redundancy resolution for the same Cartesian path $\hat{\mathbf{T}}_{\text{EE1}}(t)$ in equation (9). In the proposed algorithm, we calculate the redundancy resolution with 100 path points per second, and the parameter $m$ is set to 4000. In this path, the end effector of the manipulator will undergo an acceleration-deceleration motion tangential to a circular path with a radius of 0.1 meters, completing one full revolution. The computed path results are shown in Figure 4.

\begin{figure}[h]
	\centering
	\includegraphics[width=3.5in]{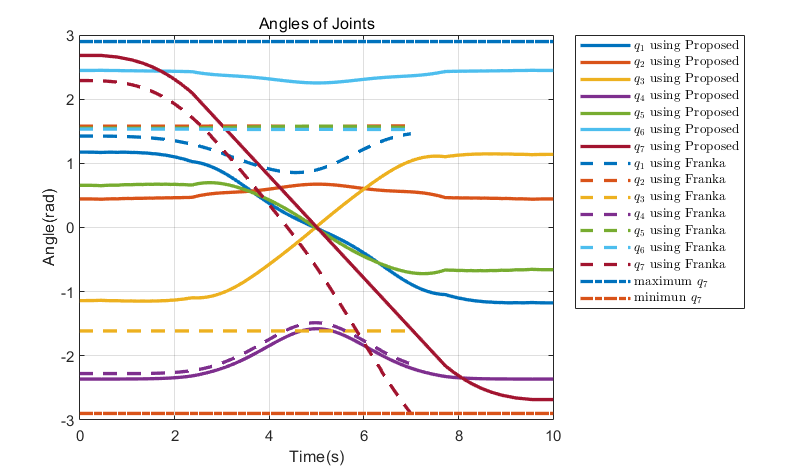}
	\caption{The computed path results of the proposed algorithm based on DP(dynamic programming) and the Franka Cartesian pose generator. The joint angle $q_7$ obtained from the Franka Cartesian pose generator reaches its lower bound during the path, leading to the inability to continue the motion and resulting in a forced termination.} 
	
	\label{fig_4}
\end{figure}

The joint angle $q_7$ obtained from the Franka Cartesian pose generator reaches its lower bound during the trajectory, leading to the inability to continue the motion and resulting in a forced termination. The proposed algorithm, on the other hand, provides a complete path in the joint space. The trajectories of the end-effector for both algorithms, plotted using Matlab, are illustrated in Figure 5. The left plot depicts the manipulator path obtained from the proposed algorithm, closely aligning with the original path. On the right, the plot shows the motion trajectory of the manipulator under the Franka Cartesian pose generator, with its start and end points indicated. It is evident that the manipulator is unable to complete the entire path seamlessly.

\begin{figure*}[ht]
	\centering
	{\includegraphics[width=3in]{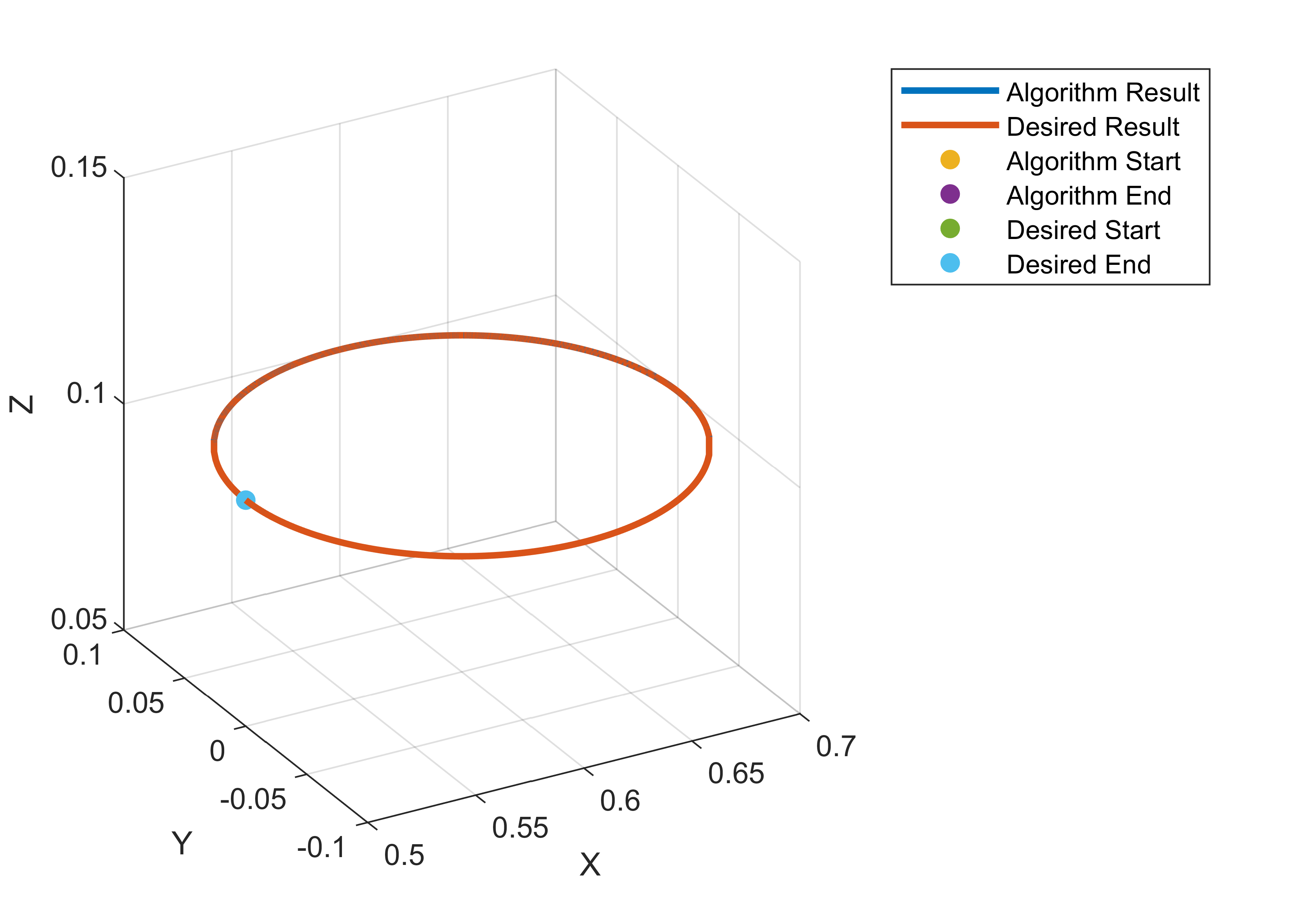}%
		\label{a}}
	\hfil
	{\includegraphics[width=3in]{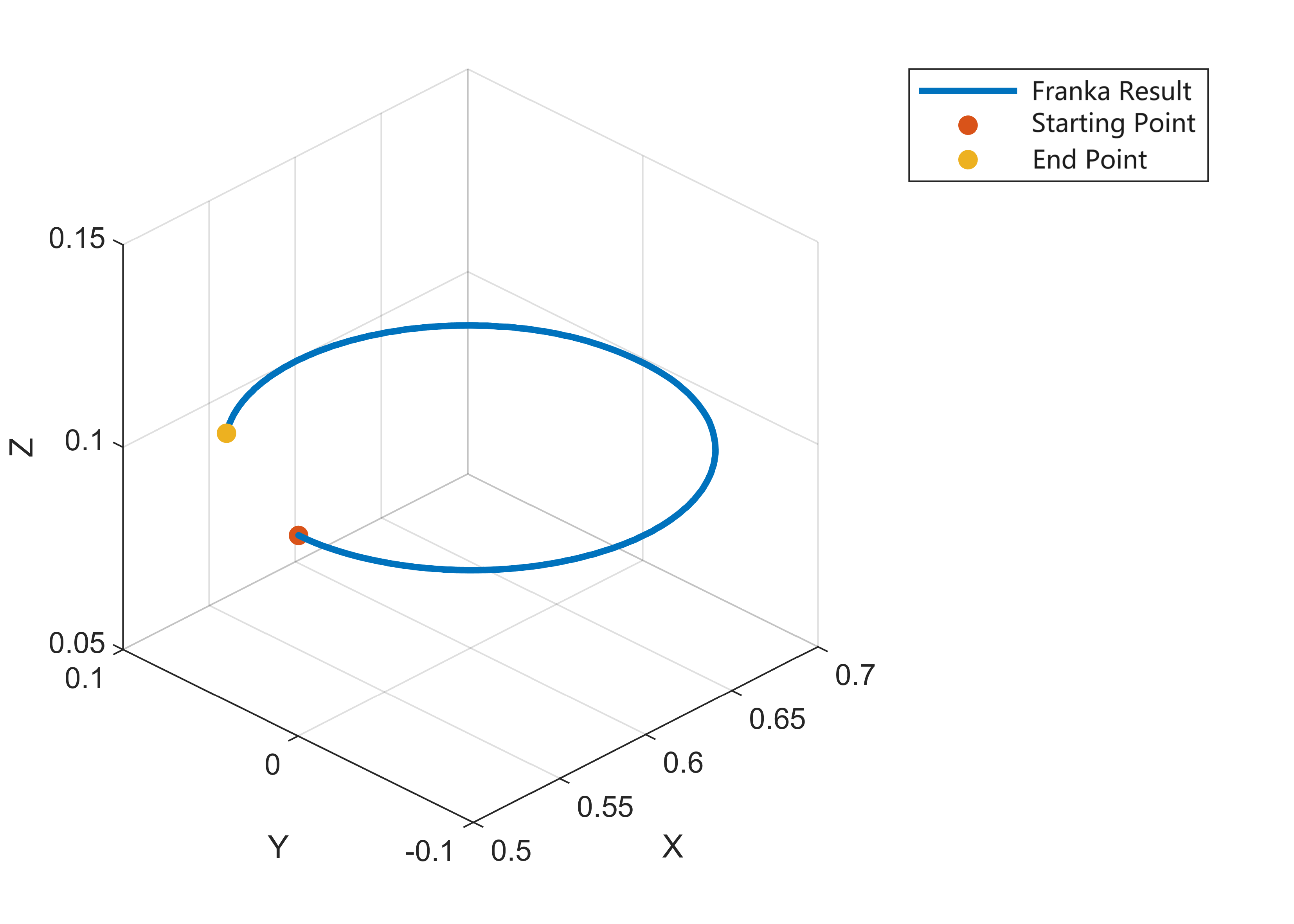}%
		\label{b}}
	\caption{The left plot depicts the manipulator path obtained from the proposed algorithm, closely aligning with the original path. On the right, the plot shows the motion trajectory of the manipulator under the Franka Cartesian pose generator, with its start and end points indicated. It is evident that the manipulator is unable to complete the entire path seamlessly.}
	\label{fig_5}
\end{figure*}

Figure 6 presents the experimental results of the actual manipulator motion based on the paths generated by the two algorithms. The top row illustrates the results obtained from the proposed algorithm, while the bottom row shows the outcomes of the Franka algorithm. Starting from the initial position, both algorithms select different poses, resulting in divergent final motion results. Under the proposed algorithm, the manipulator returns to its original pose, whereas under the Franka algorithm, the manipulator stops midway.

\begin{figure*}[ht]
	\centering
	{\includegraphics[width=6in]{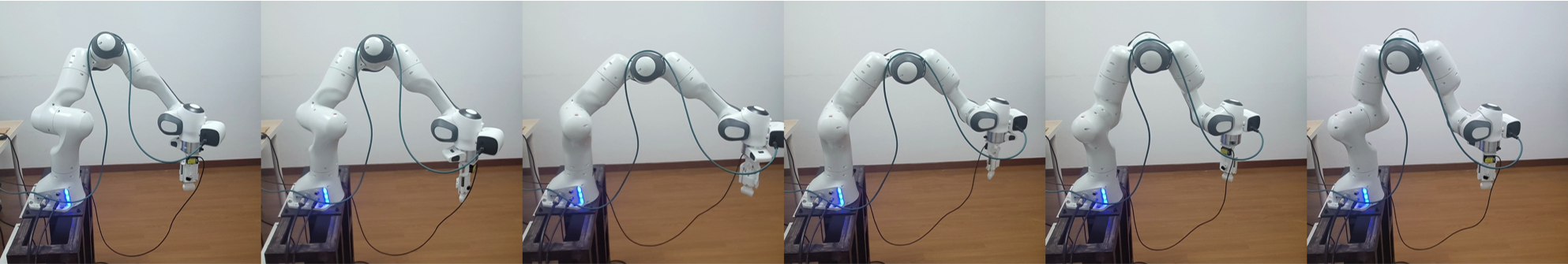}%
		\label{a}}
	\hfil
	{\includegraphics[width=6in]{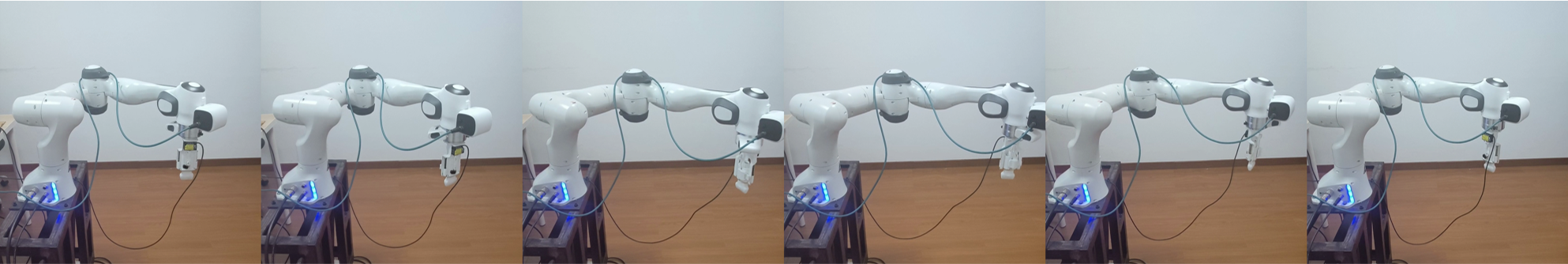}%
		\label{b}}
	\caption{The top row illustrates the results obtained from the proposed algorithm, while the bottom row shows the outcomes of the Franka algorithm. Starting from the initial position, both algorithms select different poses, resulting in divergent final motion results. Under the proposed algorithm, the manipulator returns to its original pose, whereas under the Franka algorithm, the manipulator stops midway.}
	\label{fig_6}
\end{figure*}

The utilization of dynamic programming enables the proposed algorithm to determine the joint angle values based on the poses of points along the entire path, rather than solely relying on the current position. As demonstrated in appendix B, when the value of $m$ is fixed, given the existence of a feasible path, the algorithm in this paper is guaranteed to find a globally optimal solution. In comparison to the Franka algorithm, the proposed algorithm is capable of more consistently and reliably identifying the complete path for the manipulator.

Through the aforementioned tests, it can be observed that for a given Cartesian path, obtaining feasible solutions under given constraints through traditional local methods for inverse kinematics is likely to be challenging, resulting in the manipulator being forced to stop during operation. In contrast, the proposed algorithm possesses a powerful capability to find feasible solutions. As long as feasible redundancy resolutions exist for the path, the algorithm can find a globally optimal joint space path, fully utilizing the pose information of each path point along the entire path when solving the inverse kinematics for each point. This characteristic ensures the stable and reliable operation of the manipulator. Moreover, the proposed algorithm also possesses the optimization capability for the global loss function, which cannot be achieved by local optimization algorithms.

\subsection{Error Analysis}

To verify whether the interpolated path obtained by the algorithm complies with the various constraints of the manipulator, we disregarded errors caused by actual experiments and computationally simulated the joint angles of the manipulator at various path points. Subsequently, we calculated the corresponding velocities, accelerations, and jerks at these points, and the motion parameters are normalized to the range [-1, 1], where -1 and 1 respectively correspond to the minimum and maximum values of each parameter. These normalized values were then graphically depicted, as shown in Figure 7. 

\begin{figure}[h]
	\centering
	\includegraphics[width=6in]{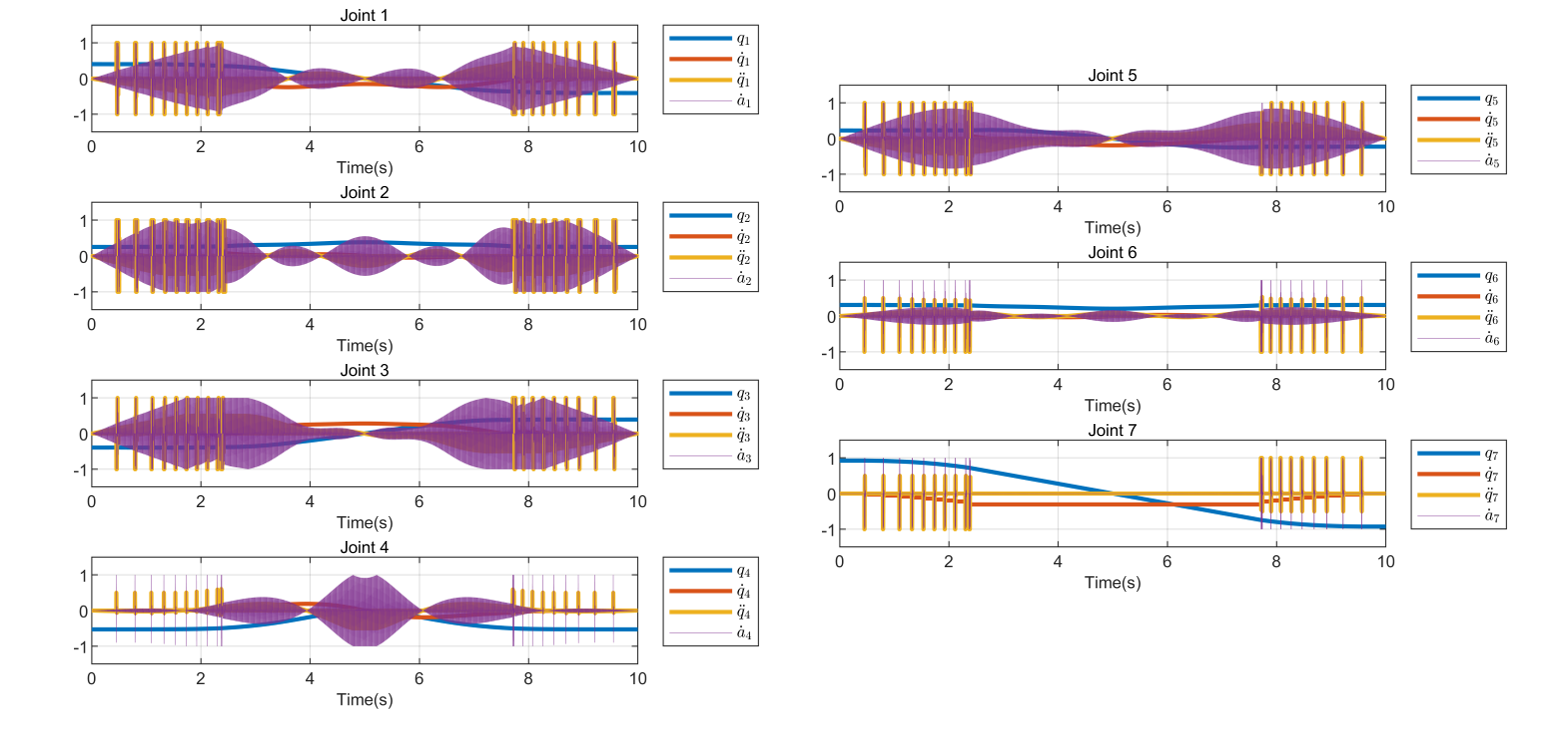}
	\caption{The angles, velocities, accelerations and jerk of each joint are normalized to the range [-1, 1], where -1 and 1 respectively correspond to the minimum and maximum values of each parameter. The figure demonstrates that the constraints on the angles, velocities, accelerations and jerks of each joint have been duly satisfied} 
	
	\label{fig_7}
\end{figure}

The figure demonstrates that the constraints on the angles, velocities, accelerations and jerks of each joint have been duly satisfied.

For the proposed algorithm, we also calculate its error with respect to the desired path. We will examine the distance between the algorithm-derived end-effector position and the expected position along the path. For linear interpolation, the more interpolation nodes there are, the smaller the error between the result and the actual value tends to be. However, in the problem under investigation, due to the joint constraints of the manipulator, the manipulator finds it challenging to reach the expected joint angles obtained through interpolation, leading to additional errors. Our compensatory algorithm aims to minimize this error as much as possible. Nevertheless, if the manipulator cannot reduce the error to zero between two adjacent sampled points, the error will accumulate. 

To assess the errors introduced by the algorithm, we conducted redundancy resolution for the test path using 100 path points per second and 10 path points per second respectively. Subsequently, we computed the error between the interpolated joint angles and the expected joint angles. The results are depicted in Figure 8.

\begin{figure}[h]
	\centering
	\includegraphics[width=3.5in]{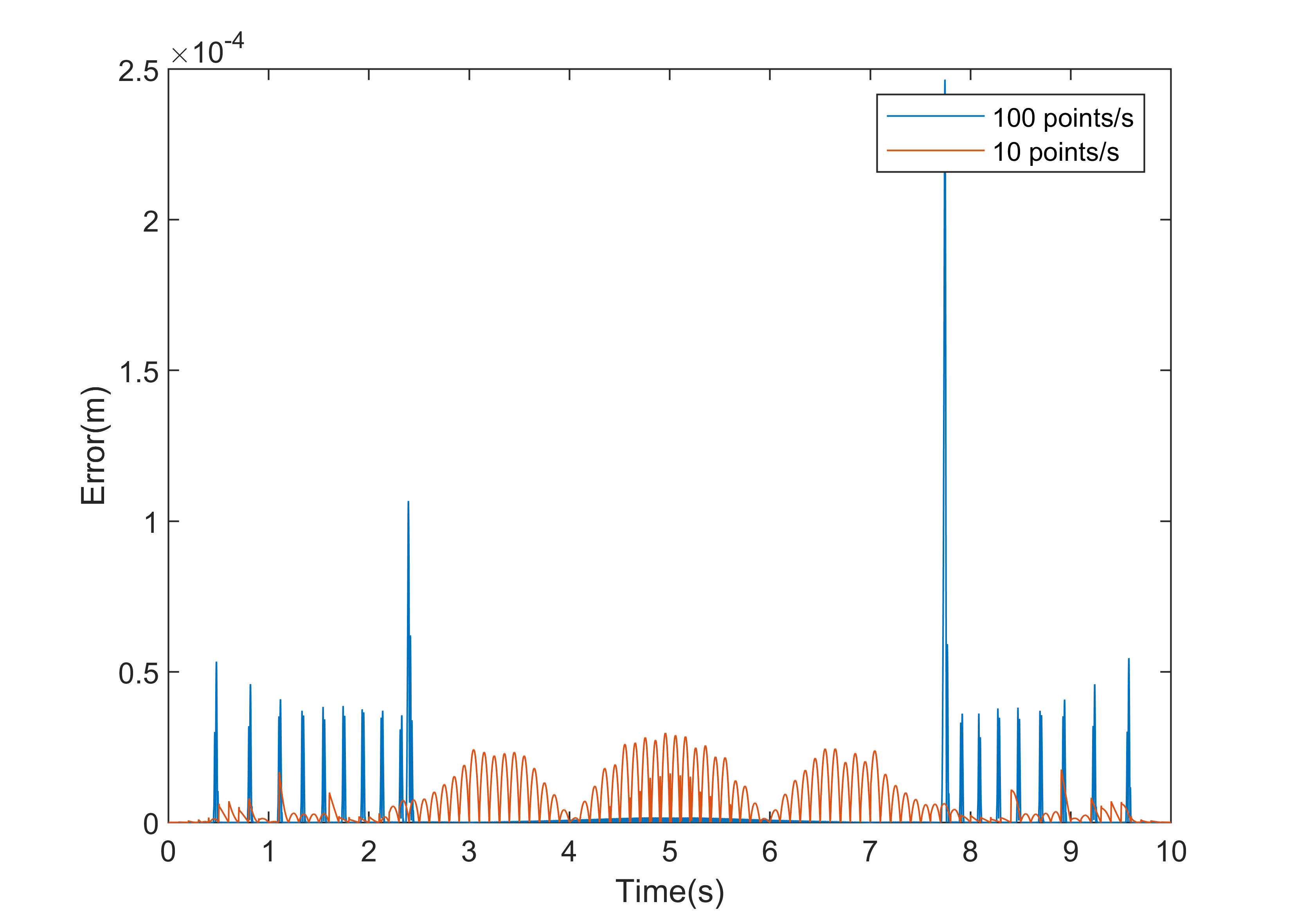}
	\caption{We conducted redundancy resolution for the test path using 100 path points per second and 10 path points per second respectively, and computed the error between the interpolated joint angles and the expected joint angles. At certain path points, the results obtained at a rate of 100 path points per second exhibited error accumulation, leading to a significant discrepancy. Through the calculation of average errors, the average error at 100 path points per second was determined to be $2.5101 \times 10^{-6} \, \text{m}$, whereas the average error at 10 path points per second was $6.7442 \times 10^{-6} \, \text{m}$. Each path demonstrates advantages and disadvantages in terms of maximum and average errors, yet overall, the errors are relatively small.} 
	
	\label{fig_8}
\end{figure}

The error between the interpolated joint angles and the expected joint angles are illustrated in Figure 8.. At certain path points, the results obtained at a rate of 100 path points per second exhibited error accumulation, leading to a significant discrepancy. Through the calculation of average errors, the average error at 100 path points per second was determined to be $2.5101 \times 10^{-6} \, \text{m}$, whereas the average error at 10 path points per second was $6.7442 \times 10^{-6} \, \text{m}$. Each path demonstrates advantages and disadvantages in terms of maximum and average errors, yet overall, the errors are relatively small.

\subsection{Test of the Starting Point Modification Algorithm}

Next, we conducted a test on the algorithm's initial point modification functionality using the path $\hat{\mathbf{T}}_{\text{EE2}}(t)$ as defined in Equation (10). Initially, we computed its redundancy resolution using Algorithm 1, resulting in the joint angles depicted in Figure 9. 

\begin{figure}[t]
	\centering
	\includegraphics[width=3.2in]{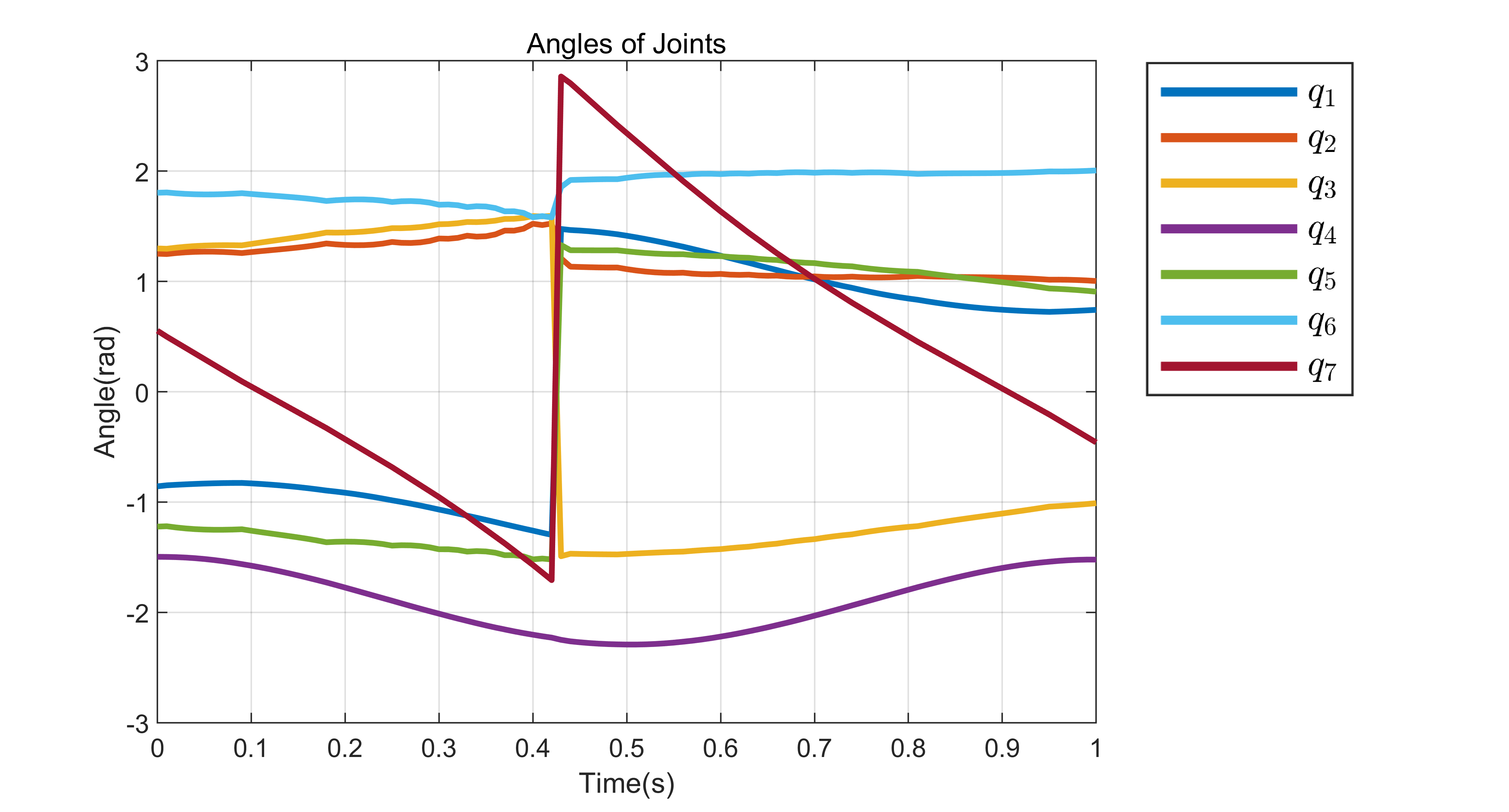}
	\caption{The redundancy resolution of $\hat{\mathbf{T}}_{\text{EE2}}(t)$ computed with Algorithm 1. It is evident from the plot that the curves representing the joint variations over time exhibit a distinct discontinuity, indicating the presence of discontinuities in the motion.} 
	
	\label{fig_9}
\end{figure}	

It is evident from the plot that the curves representing the joint variations over time exhibit a distinct discontinuity, indicating the presence of discontinuities in the motion.

Following this, we employed Algorithm 2 to compute $\hat{\mathbf{T}}_{\text{EE2}}(t)$. The algorithm successfully identified a new starting point that reduced the number of discontinuities in the motion. Subsequently, upon altering the starting point, the joint angles of the manipulator, as depicted in Figure 10, no longer exhibit any discontinuities over time.	
\begin{figure}[t]
	\centering
	\includegraphics[width=3.2in]{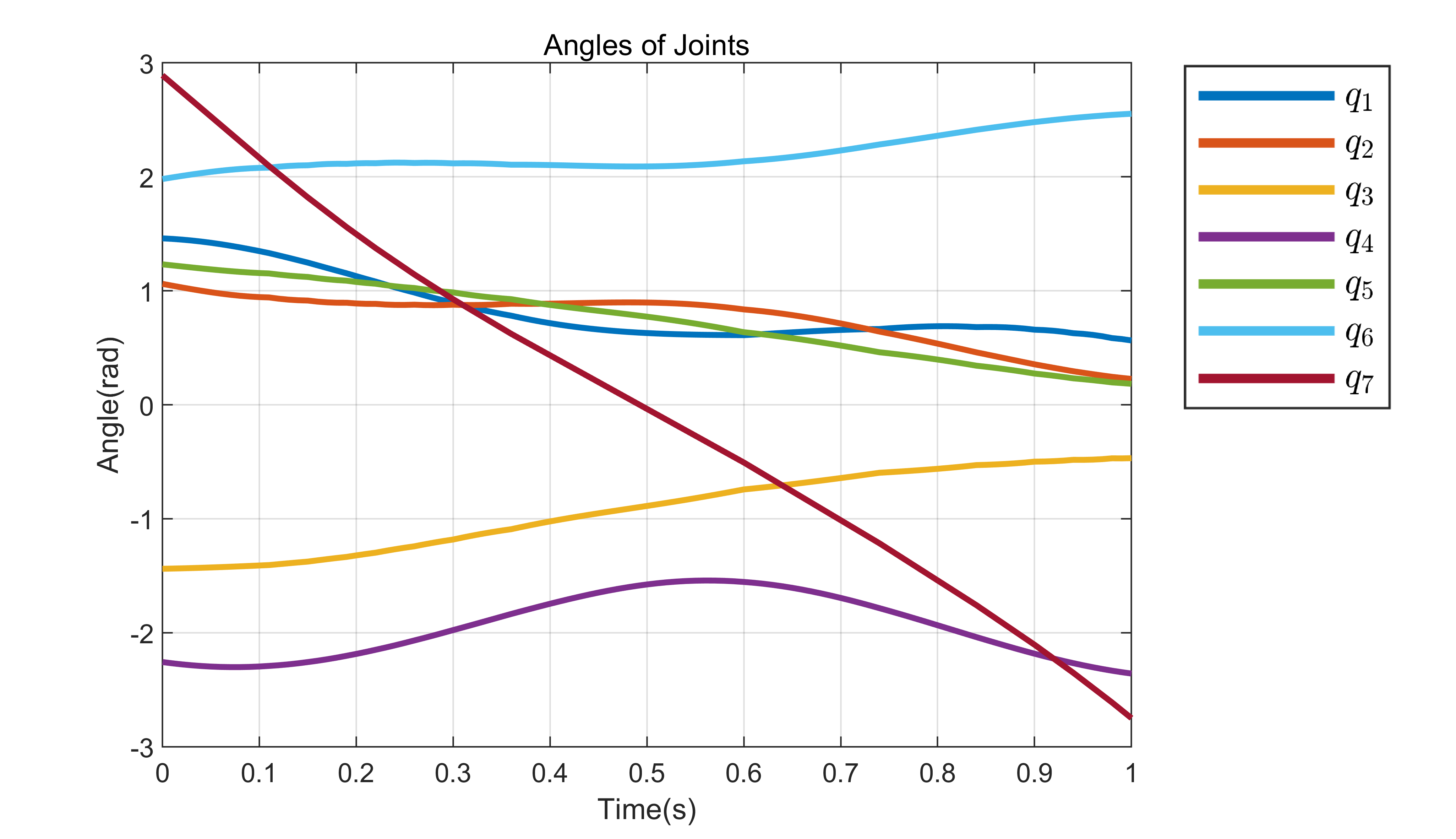}
	\caption{We employed Algorithm 2 to compute $\hat{\mathbf{T}}_{\text{EE2}}(t)$. The algorithm successfully identified a new starting point that reduced the number of discontinuities in the motion. Upon altering the starting point, the joint angles of the manipulator, as depicted in this figure, no longer exhibit any discontinuities over time.	} 
	
	\label{fig_10}
\end{figure}	

Following this, we plot the manipulator poses corresponding to the results obtained from both algorithms using Matlab, and compare them with the desired path, as shown in Figure 11.
\begin{figure}[ht]
	\centering
	{\includegraphics[width=3in]{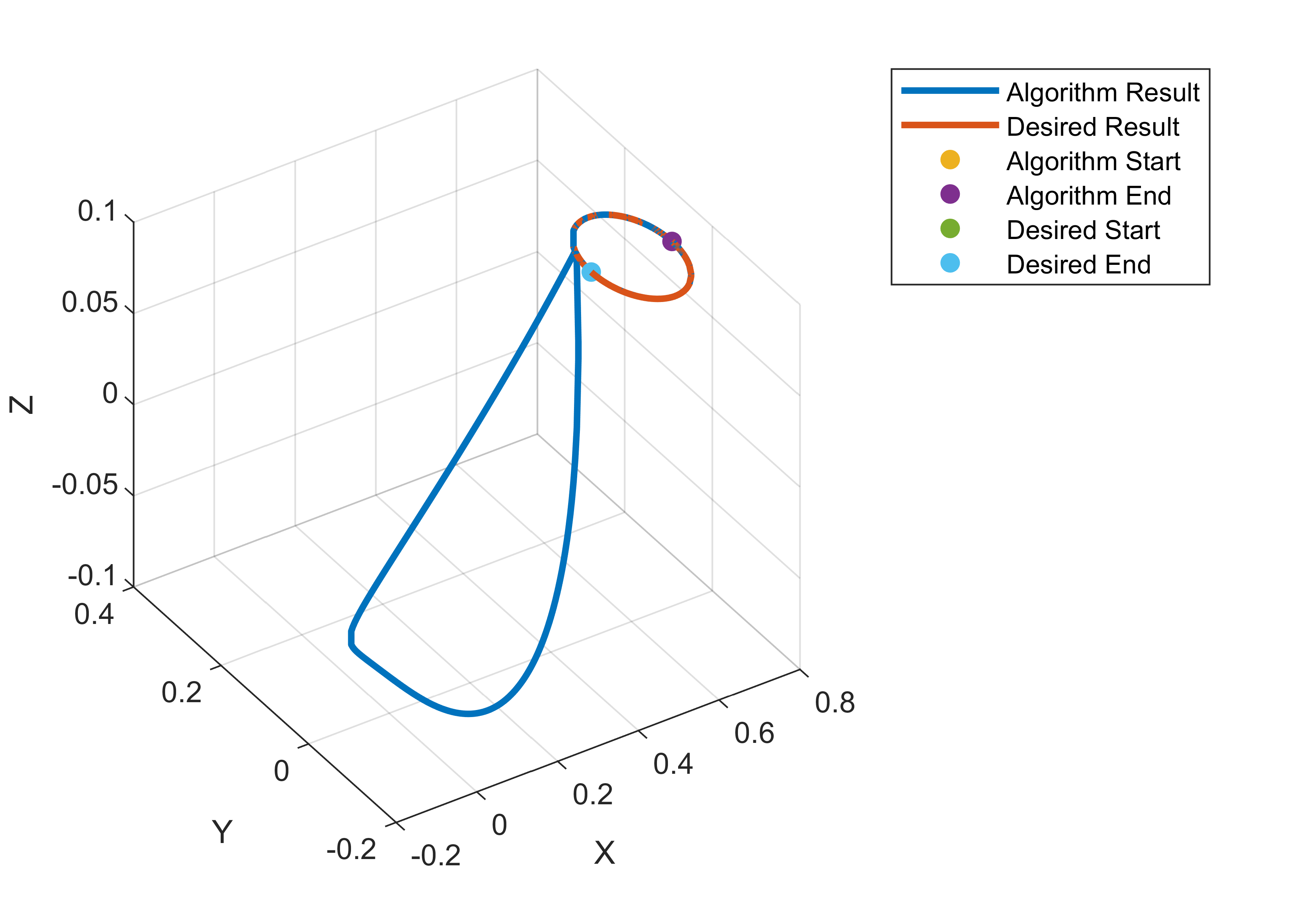}%
		\label{a}}
	\hfil
	{\includegraphics[width=3in]{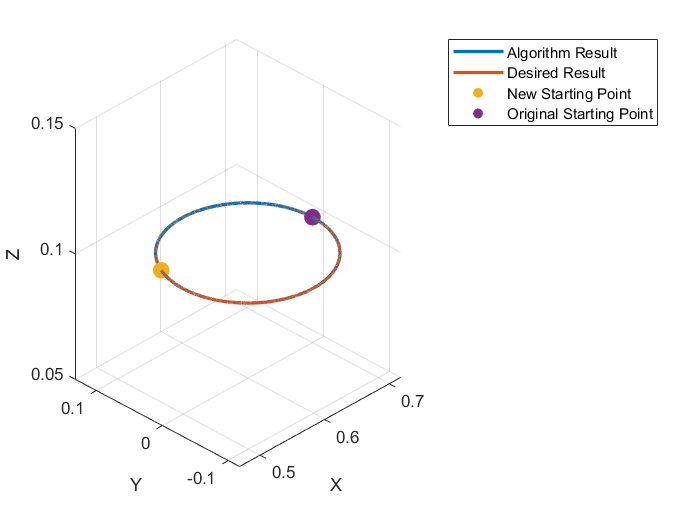}%
		\label{b}}
	\caption{The manipulator poses corresponding to the results obtained from both algorithms.The upper image corresponds to the outcome of Algorithm 1, revealing that the manipulator deviates significantly from the expected path midway, circling in a different direction before continuing along the intended trajectory. On the other hand, the lower image represents the outcome of Algorithm 2. After modifying the starting point, the manipulator can traverse the expected path continuously without interruption.}
	\label{fig_11}
\end{figure}

The upper image corresponds to the outcome of Algorithm 1, revealing that the manipulator deviates significantly from the expected path midway, circling in a different direction before continuing along the intended trajectory. This large deviation represents a discontinuous motion of the manipulator, involving an adjustment of joint angles. On the other hand, the lower image represents the outcome of Algorithm 2. After modifying the starting point, the manipulator can traverse the expected path continuously without interruption.

\section{Conclusion}

This paper proposes a redundancy resolution algorithm for a 7-DOF manipulator based on dynamic programming. The dynamic programming approach allows for comprehensive consideration of the entire path. By leveraging this methodology, the algorithm optimizes the joint angles throughout the entire path. The algorithm is capable of finding a globally optimal solution for the path that minimizes the cost function and ensures the path has the fewest breakpoints, thereby enhancing the operational efficiency of the manipulator. The proposed algorithm innovatively determines the minimum number of breakpoints required for the manipulator to traverse along a path without necessitating additional analysis, and optimizes the breakpoints and the starting positions of circular paths, constituting the primary innovation of this study.

\section*{Acknowledgment}

The author would like to thank Siyuan Jiang and Chaowei Chen from Zhejiang Demetics Medical Technology Co., Ltd. for providing necessary guidance, resources and facilities. This work would not have been possible without their collective support.  

This work was supported by The National Natural Science Foundation(Nos.12090020, 12090025) and the Zhejiang Provincial Science and Technology Program(2022C03113).

\bibliographystyle{elsarticle-num}

\bibliography{cas-refs}



\end{document}